\documentclass[runningheads]{llncs}

\usepackage{eccv}

\usepackage[width=122mm,left=12mm,paperwidth=146mm,height=193mm,top=12mm,paperheight=217mm]{geometry}

\usepackage{eccvabbrv}
\usepackage{graphicx}
\usepackage{booktabs}
\usepackage{multirow}
\usepackage[table]{xcolor}
\usepackage{svg}
\usepackage[accsupp]{axessibility}
\usepackage{amsmath}

\usepackage{algorithm}
\usepackage{algpseudocode}

\usepackage{tikz}
\usepackage{pgfplots}
\pgfplotsset{compat=1.18}

\usepackage[breaklinks,colorlinks,citecolor=eccvblue]{hyperref}

\usepackage{orcidlink}

\begin{document}

\title{Matryoshka Gaussian Splatting}

\author{Zhilin Guo\inst{1}\orcidlink{0000-0002-7660-3102}, 
Boqiao Zhang\inst{1}\orcidlink{0009-0000-0745-2438}, 
Hakan Aktas\inst{1}\orcidlink{0009-0009-4796-4281}, 
\\
Kyle Fogarty\inst{1}\orcidlink{0000-0002-1888-4006},
Jeffrey Hu\inst{1}\orcidlink{0009-0003-4400-0862}, 
Nursena Koprucu Aslan\inst{1}\orcidlink{0009-0007-2971-8840}, 
Wenzhao Li\inst{1}\orcidlink{0009-0007-0265-463X}, 
Canberk Baykal\inst{1}\orcidlink{0000-0002-0249-5858}, 
Albert Miao\inst{1}\orcidlink{0009-0002-4465-9563}, 
Josef Bengtson\inst{2}\orcidlink{0000-0002-4321-491X}, 
Chenliang Zhou\inst{1}\orcidlink{0009-0004-4687-6945}, 
\\
Weihao Xia\inst{1}\thanks{Corresponding author: wx258@cam.ac.uk}\orcidlink{0000-0003-0087-3525}, 
Cristina Nader Vasconcelos\inst{3}\orcidlink{0000-0003-2112-4806}, 
Cengiz Oztireli\inst{1,3}\orcidlink{0000-0002-4700-2236}
}

\authorrunning{Z.~Guo et al.}

\institute{$^{1}$University of Cambridge  \ \ \ \ 
$^{2}$Chalmers University of Technology \ \ \ \ 
$^{3}$Google
\\
\vspace{3pt}
{\small 
\url{https://ZhilinGuo.github.io/MGS}}
}

\titlerunning{Matryoshka Gaussian Splatting}

\maketitle

\begin{abstract}
  The ability to render scenes at adjustable fidelity from a single model, known as level of detail (LoD), is crucial for practical deployment of 3D Gaussian Splatting (3DGS).
Existing discrete LoD methods expose only a limited set of operating points, while concurrent continuous LoD approaches enable smoother scaling but often suffer noticeable quality degradation at full capacity, making LoD a costly design decision.
We introduce \emph{Matryoshka Gaussian Splatting (MGS)}, 
a training framework that enables continuous LoD for standard 3DGS pipelines without sacrificing full-capacity rendering quality.
MGS learns a single ordered set of Gaussians such that rendering any prefix, the first $k$ splats, produces a coherent reconstruction whose fidelity improves smoothly with increasing budget.
Our key idea is \emph{stochastic budget training}: each iteration samples a random splat budget and optimises both the corresponding prefix and the full set. This strategy requires only two forward passes and introduces no architectural modifications.
Experiments across four benchmarks and six baselines show that MGS matches the full-capacity performance of its backbone while enabling a continuous speed–quality trade-off from a single model.
Extensive ablations on ordering strategies, training objectives, and model capacity further validate the designs.

\end{abstract}

\section{Introduction}
\label{sec:introduction}

\begin{figure*}[t]
  \centering
    \includegraphics[width=1.0\linewidth]{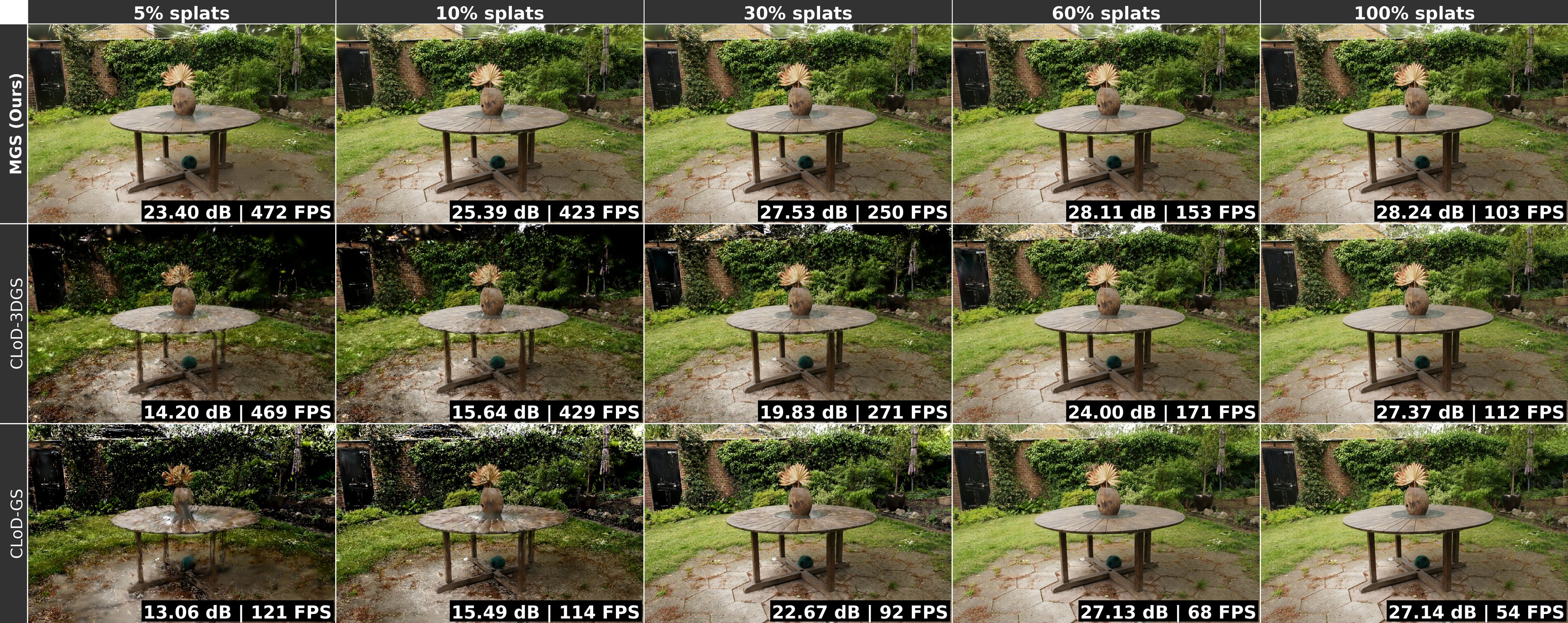}
    \caption{
    Continuous LoD need not sacrifice full-capacity quality to enable budget trade-off.
    Our method, MGS (top), learns an ordered set of Gaussian primitives
    whose prefixes yield coherent reconstructions at any splat budget.
    Compared to CLoD-3DGS~\cite{CLoD-3DGS-milef2025learning} (mid) and
    CLoD-GS~\cite{CLoD-GS-cheng2025clod} (bot),
    MGS achieves the highest fidelity at every operating point
    with quality degrading gracefully under budget reduction.  Scene: \emph{Garden} \cite{MipNeRF360-barron2022mip}.
    }
  \label{fig:teaser}
\end{figure*}

Real-time neural rendering is governed by a fundamental tension between scene fidelity and the computational budget available \cite{NeRF-mildenhall2020nerf, KiloNeRF}.
This budget varies by orders of magnitude across the hardware spectrum, from high-end GPU workstations to mobile devices and mixed-reality headsets\cite{compact_neural_radiance, Wang2023Foveated}; and fluctuates dynamically at runtime by viewpoint and scene complexity \cite{funkhouser1993adaptive, Zhang2018OntheFlyPR}.
Level of detail (LoD) techniques address this tension by scaling the rendered representation to match available resources and have long been a cornerstone of interactive graphics~\cite{luebke2002level}.
The dominant paradigm, discrete LoD, precomputes a set of quality levels and switches between them at runtime~\cite{clark1976hierarchical}.
A coarse set of fixed levels, however, cannot smoothly track a budget that shifts continuously with scene content and viewpoint, and the abrupt transitions between levels produce visible pop-in and pop-out artifacts~\cite{funkhouser1993adaptive, giegl2007unpopping}.

3D Gaussian Splatting (3DGS)~\cite{3DGS-kerbl20233d} achieves photorealistic novel view synthesis by rasterizing millions of anisotropic Gaussian primitives 
at real-time frame rates, with computational cost scaling directly with the number of primitives~\cite{fridovich2022plenoxels}.
In principle, this primitive-based nature of the representation offers continuous budget control, since omitting any subset of splats yields an immediate speedup \cite{3DGS-kerbl20233d, PUP-3DGS-hanson2025pup}.
Yet a conventionally trained 3DGS model has no ordering among its primitives, so quality collapses rapidly as splats are removed \cite{lee2026safeguardgs}.

Existing approaches attempt to introduce LoD to 3DGS through several strategies.
Discrete LoD methods~\cite{H3DGS-kerbl2024hierarchical,Octree-GS-ren2024octree,FLoD-seo2024flod,LODGE-kulhanek2025lodge} build hierarchical structures over Gaussian primitives, exposing a limited number of quality levels but requiring auxiliary index structures and offering only coarse budget granularity.
Compression and pruning pipelines~\cite{LightGaussian-fan2024lightgaussian,MaskGaussian-liu2025maskgaussian,FlexGaussian-tian2025flexgaussian} can approximate multi-budget behaviour, but each operating point is typically obtained independently without ensuring that subsets are nested or globally coherent.
Concurrent continuous LoD methods~\cite{CLoD-GS-cheng2025clod,CLoD-3DGS-milef2025learning} enable smoother scaling within a single model, but often incur substantial quality degradation at full capacity.
As a result, adopting LoD in 3DGS remains a costly design choice that often sacrifices reconstruction quality.

In this work, we introduce \emph{Matryoshka Gaussian Splatting} (MGS), a training framework that enables continuous LoD control without degrading full-capacity quality \cref{fig:teaser}.
Where Matryoshka Representation Learning~\cite{MRL-kusupati2022matryoshka} nests several embedding dimensions, MGS transfers this principle to scene primitives at arbitrary granularity.
Concretely, MGS learns an ordered set of Gaussian primitives such that any prefix (the first $k$ splats) forms a coherent scene representation, with reconstruction fidelity improving smoothly as $k$ increases.
Rendering at different budgets is then achieved by truncating the ordered sequence, producing a continuous spectrum of quality-speed operating points from a single model.

The key mechanism is \emph{stochastic budget training}: each iteration samples a random splat budget uniformly and optimises both the corresponding prefix and the full set, requiring only two forward passes each training step.
Because this procedure modifies only the training objective and not the model architecture, it integrates into most existing 3DGS pipelines with minimal implementation effort.
At deployment, one adjusts $k$ to match available resources---no per-budget retraining, no auxiliary data structures.

Extensive experiments across four benchmarks~\cite{MipNeRF360-barron2022mip, TanksAndTemples-knapitsch2017tanks, DeepBlending-hedman2018deep, BungeeNeRF-xiangli2022bungeenerf} and six baselines~\cite{CLoD-GS-cheng2025clod, CLoD-3DGS-milef2025learning, Octree-GS-ren2024octree, MaskGaussian-liu2025maskgaussian, FlexGaussian-tian2025flexgaussian, H3DGS-kerbl2024hierarchical} spanning discrete LoD and continuous LoD methods demonstrate that MGS matches the full-capacity reconstruction quality of its backbone, while enabling a continuous quality-speed frontier from a single model.

Our contributions are summarised as follows:
\begin{itemize}
  \item We introduce \textit{Matryoshka Gaussian Splatting} (MGS), which learns an nested Gaussian primitive representation where any prefix yields a coherent scene reconstruction, enabling continuous LoD control.
  \item We propose \textit{stochastic budget training}, a simple yet effective model-agnostic  training procedure that optimises across a continuous budget range with only two renders per iteration. %
  \item We provide extensive evaluation on four benchmarks against six baselines, demonstrating state-of-the-art quality-speed performance, with ablations on importance scoring, budget sampling, and training objective.
\end{itemize}

\begin{figure*}[t]
  \centering
  \includegraphics[width=0.8\linewidth]{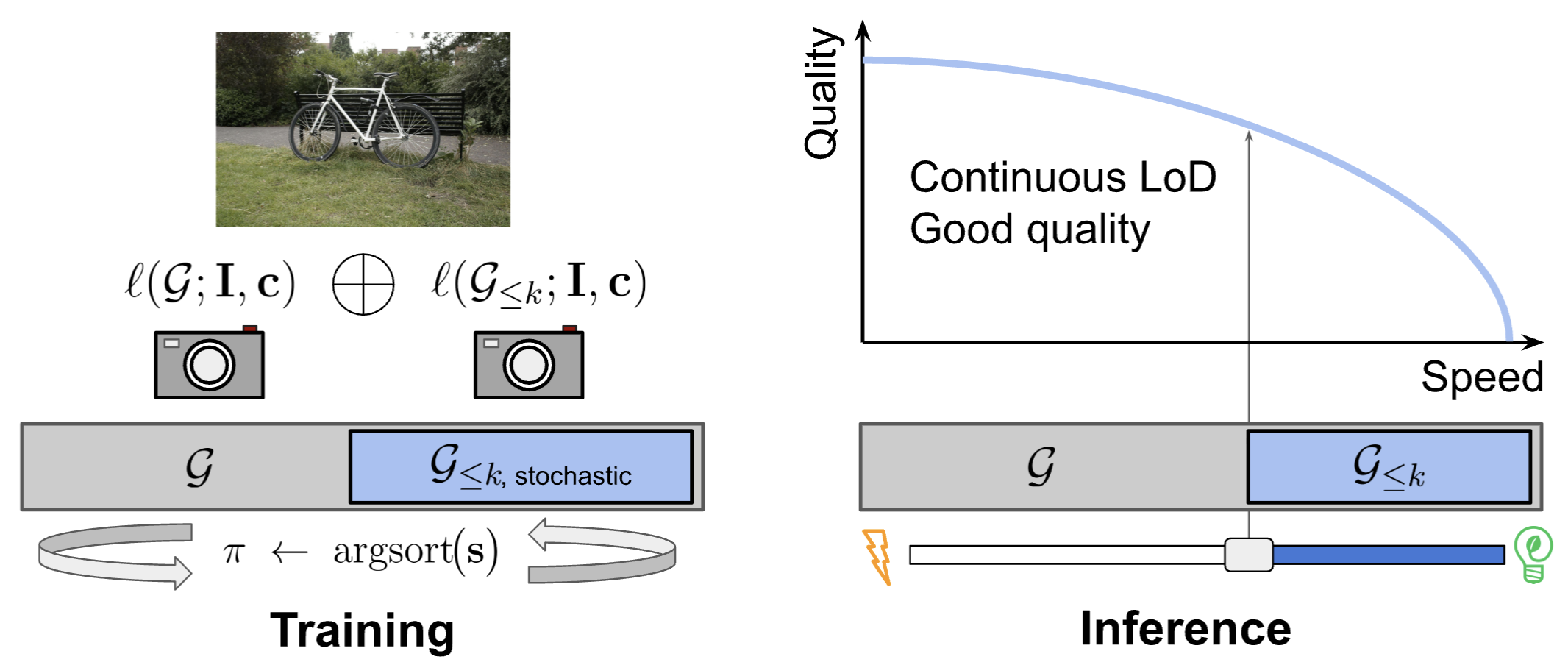}
  \caption{Framework of Matryoshka Gaussian Splatting.}
  \label{fig:overview}
\end{figure*}
\section{Related Work}
\label{sec:related-work}

\subsection{3D Gaussian Splatting}

3D Gaussian Splatting (3DGS)~\cite{3DGS-kerbl20233d} replaces implicit volumetric representations in Neural Radiance Fields~\cite{NeRF-mildenhall2020nerf} with explicit anisotropic Gaussian primitives rendered via differentiable rasterization. This formulation enables photorealistic novel view synthesis at real-time frame rates, with rendering cost scaling directly with the number of primitives, making primitive count a natural axis for controlling computational budget.
Subsequent work~\cite{3DGS-MCMC-kheradmand20243d,Mini-Splatting-fang2024mini,Mip-Splatting-yu2024mip,MS-3DGS-yan2024multi} has extended 3DGS along several directions.
For capacity control, the original densification heuristics offer no direct control over the final primitive count.
3DGS-MCMC~\cite{3DGS-MCMC-kheradmand20243d} addresses this via Langevin-dynamics sampling under a fixed budget. Mini-Splatting~\cite{Mini-Splatting-fang2024mini} further improves densification under constrained capacities.
For multi-scale rendering, Mip-Splatting~\cite{Mip-Splatting-yu2024mip} and Multi-Scale 3DGS~\cite{MS-3DGS-yan2024multi} introduce anti-aliasing filters that improve quality when viewed at varying observation scales.
Our method MGS builds upon these advances by learning a primitive
ordering that maintains coherent reconstructions across a continuous
range of rendering budgets.

\subsection{Level-of-Detail Rendering} %

Level-of-detail (LoD) rendering enables adaptive control of rendering cost by varying scene complexity according to available computational budgets.
Discrete LoD~\cite{H3DGS-kerbl2024hierarchical,Scaffold-GS-lu2024scaffold,Octree-GS-ren2024octree,FLoD-seo2024flod,LODGE-kulhanek2025lodge,shen2025lod} and progressive streaming~\cite{PCGS-chen2025pcgs,LapisGS-shi2025lapisgs} methods construct hierarchical or layered structures over Gaussians to expose multiple quality levels.
However, these methods require auxiliary index structures and typically provide only a small number of discrete operating points \cite{Octree-GS-ren2024octree, H3DGS-kerbl2024hierarchical}.
Compression and pruning pipelines~\cite{LightGaussian-fan2024lightgaussian,MaskGaussian-liu2025maskgaussian,FlexGaussian-tian2025flexgaussian} can approximate multi-budget behaviour by applying importance-based pruning or quantization at varying targets, but each operating point is typically obtained independently without ensuring nested or globally consistent primitive subsets \cite{CompGS-navaneet2024compgs, Reduced-3DGS-papantonakis2024reducing}.
Concurrent continuous LoD methods~\cite{CLoD-GS-cheng2025clod,CLoD-3DGS-milef2025learning} target smoother budget scaling within a single model, either by learning view-dependent opacity decay \cite{View_dependent_PM_refinement} or by training on random primitive subsets.
However, both incur noticeable quality degradation at full capacity and rapid quality collapse at reduced budgets, making continuous LoD a costly design choice.
In contrast, MGS learns a single ordered set of Gaussian primitives such that \emph{every} prefix yields a coherent reconstruction.
This produces a dense continuum of valid budgets while still closely matching the backbone's full-capacity quality.

\subsection{Nested (Matryoshka) Representations}

Nested representations learn ordered structures in which every prefix of the representation remains independently usable. 
This idea was introduced by nested dropout~\cite{nested-dropout-rippel2014learning} and scaled to high-dimensional embeddings by Matryoshka Representation Learning (MRL)~\cite{MRL-kusupati2022matryoshka}.
Related ideas of packing multiple capacity levels into a single model have also been explored in slimmable networks~\cite{Slimmable-yu2019slimmable,OFA-cai2020once}, which train neural networks that operate at different channel widths, although these methods typically require separate forward passes for each width configuration. 
In classical graphics, progressive meshes~\cite{PM-hoppe1996progressive} provide an analogous concept by representing geometry as an ordered sequence of refinement operations that enables continuous mesh simplification.
MGS transfers the nested representation principle to Gaussian scene primitives to enable continuous control over rendering budgets through ordered primitive prefixes within a single model.

\section{Method}
\label{sec:method}

Our MGS consists of two components: (i)~an \emph{ordered Gaussian representation} where importance-ranked prefixes serve as nested scene representations at different fidelity levels (\cref{sec:mgs_representation}), and (ii)~a \emph{stochastic training procedure} that efficiently optimises across all prefix lengths using a single random prefix and the full set per iteration (\cref{sec:mgs_training}).
We first review the necessary background in \cref{sec:prelim}.

\subsection{Preliminaries on 3DGS}
\label{sec:prelim}

3D Gaussian Splatting (3DGS)~\cite{3DGS-kerbl20233d} represents a scene as a set of $N$ anisotropic 3D Gaussians $\mathcal{G}=\{g_i\}_{i=1}^{N}$.
Each Gaussian $g_i$ carries a mean $\boldsymbol{\mu}_i\in\mathbb{R}^3$, a covariance $\boldsymbol{\Sigma}_i$, an opacity $\sigma_i\in[0,1]$, and view-dependent color parameters (\eg, spherical harmonics).
Given a camera $\mathbf{c}$, a pixel color is obtained via front-to-back $\alpha$-compositing of depth-sorted Gaussians:
\begin{equation}
\label{eq:alpha_comp}
\hat{\mathbf{I}}(\mathbf{u})
=
\sum_{i=1}^{N} \mathbf{C}_i(\mathbf{u})\,\alpha_i(\mathbf{u})
\prod_{j=1}^{i-1}\bigl(1-\alpha_j(\mathbf{u})\bigr),
\end{equation}
where $\mathbf{u}$ is a pixel coordinate, $\mathbf{C}_i(\mathbf{u})$ the view-dependent color, and $\alpha_i(\mathbf{u})$ the screen-space opacity obtained by evaluating the projected 2D Gaussian at $\mathbf{u}$.
Given posed training images $\mathcal{D}=\{(\mathbf{I}_n,\mathbf{c}_n)\}_{n=1}^{M}$, the model minimises a per-image reconstruction loss:
\begin{equation}
\label{eq:rec_loss}
\ell(\mathcal{G};\mathbf{I},\mathbf{c}) :=
(1-\lambda)\,\bigl\|\mathcal{R}(\mathcal{G};\mathbf{c})-\mathbf{I}\bigr\|_1
\;+\;
\lambda\,\mathcal{L}_{\mathrm{D\text{-}SSIM}}\!\bigl(\mathcal{R}(\mathcal{G};\mathbf{c}),\mathbf{I}\bigr),
\end{equation}
where $\mathcal{R}$ denotes the differentiable splatting renderer and $\lambda$ is the mixing weight.

\subsection{Matryoshka Gaussian Splatting}
\label{sec:mgs_representation}

We construct a nested Gaussian representation that enables continuous control over rendering budgets from a single model. 
Specifically, we (i) define an importance score to rank Gaussian primitives, (ii) organize them into an ordered prefix representation that supports variable-budget rendering, and (iii) adopt an 
capacity control mechanism to maintain a fixed Gaussian budget during training.

\subsubsection{Importance Score.}

To construct a nested Gaussian representation, we assign each Gaussian
primitive $g_i$ a scalar score $s(g_i)$ based on a per-primitive
property.
Gaussians are then sorted by this score so that the most important primitives appear first in the sequence.

The MGS formulation is agnostic to the choice of $s(g)$; any statistic that induces a meaningful importance ordering is admissible.
Empirically, we find that sorting Gaussians by opacity in descending order yields stable behaviour across budgets.
This is likely because opacity reflects the visibility and radiance contribution of each primitive, making it a natural criterion for ordering primitives so that early prefixes already capture the dominant scene structure.
Therefore, we use primitive's opacity as importance score in implementation:
\begin{equation}
\label{eq:importance}
s(g_i) = \sigma_i ,
\end{equation}
where $\sigma_i$ denotes the opacity parameter of Gaussian $g_i$. 
Alternative scoring criteria are evaluated in ablation studies.

\subsubsection{Nested Primitive Representation.}
Let $\pi$ be a permutation of $\{1,\dots,N\}$ that ranks Gaussians by a scalar importance score $s(g)$ in non-increasing order:
\begin{equation}
\label{eq:ordering}
s\!\bigl(g_{\pi(1)}\bigr)\ge s\!\bigl(g_{\pi(2)}\bigr)\ge \cdots \ge s\!\bigl(g_{\pi(N)}\bigr).
\end{equation}
The $k$-prefix is the subset of the $k$ highest-ranked Gaussians:
\begin{equation}
\label{eq:prefix}
\mathcal{G}_{\le k} :=
\{g_{\pi(1)},\dots,g_{\pi(k)}\},
\qquad k\in\{1,\dots,N\}.
\end{equation}
Rendering with budget $k$ produces $\hat{\mathbf{I}}_{\le k}=\mathcal{R}(\mathcal{G}_{\le k};\mathbf{c})$.
Because prefixes are nested ($\mathcal{G}_{\le k}\subset\mathcal{G}_{\le k'}$ for $k<k'$) and rasterization cost scales with the Gaussian count, varying $k$ traces a continuous quality-speed curve from a single trained model without any per-budget retraining or model switching.

\subsection{Stochastic Budget Training}
\label{sec:mgs_training}

Training a single representation to perform well at every budget level is the central challenge of LoD.
Optimizing all $N$ possible budgets per step would be prohibitively expensive.
We instead propose a stochastic procedure that covers a dense continuum of prefix sizes with only two forward passes per iteration.

\subsubsection{Budget Sampling.}
At each training step we draw a random keep ratio and compute the corresponding prefix size $k$ such that:
\begin{equation}
\label{eq:keep_ratio}
k = \left\lceil r\,N \right\rceil, \qquad r \sim \mathrm{Unif}(r_{\min},\,1),
\end{equation}
where $r_{\min}\in(0,1]$ is the smallest prefix fraction seen during training; $r_{\min}=1$ recovers standard 3DGS training.
Uniform sampling ensures that every budget in $[\lceil r_{\min}N\rceil,\,N]$ is visited with equal probability over training, yielding unbiased coverage of the full budget spectrum.

\subsubsection{Training Objective.}
At each step, we sample a training view $(\mathbf{I},\mathbf{c})\sim\mathcal{D}$ and prefix size $k=\lceil rN\rceil$, render both the prefix and the full set, and minimise:
\begin{equation}
\label{eq:mgs_objective}
\ell_{\mathrm{MGS}} :=
\ell(\mathcal{G}_{\le k};\mathbf{I},\mathbf{c})
\;+\;
\gamma\,\ell(\mathcal{G}_{\le N};\mathbf{I},\mathbf{c}),
\end{equation}
where $\gamma\ge 0$ balances the prefix and full-set terms.
The prefix term pressures the model to produce strong reconstructions from partial subsets, while the full-set term anchors full-quality performance.
Over training, this stochastic procedure estimates the expected multi-budget objective across all budget fractions $r\in[r_{\min},1]$ and all training views, with prefix membership determined by the ordering in \cref{eq:ordering}.
Each step incurs exactly two renders regardless of $N$.

\subsubsection{Dynamic Reordering.}
Because gradient updates modify the Gaussian parameters at every step, the importance scores $s(g_i)$ evolve throughout training.
After each training iteration, we recompute the permutation so that the ordering in \cref{eq:ordering} holds under the current parameters:
\begin{equation}
\label{eq:reorder}
\pi \;\leftarrow\; \operatorname{argsort}\!\bigl(\mathbf{s}\bigr), \qquad \mathbf{s} = \bigl(s(g_1),\dots,s(g_N)\bigr),
\end{equation}
in non-increasing order.
This ensures every prefix $\mathcal{G}_{\le k}$ always contains the $k$ most important primitives under the current parameters.
\section{Experiments}
\label{sec:experiments}

\subsection{Experimental Setup}

\subsubsection{Benchmarks.}
We evaluate MGS on four standard 3DGS benchmarks:
{MipNeRF 360}~\cite{MipNeRF360-barron2022mip},
{Tanks \& Temples}~\cite{TanksAndTemples-knapitsch2017tanks},
{Deep Blending}~\cite{DeepBlending-hedman2018deep},
and {BungeeNeRF}~\cite{BungeeNeRF-xiangli2022bungeenerf}.
For all datasets, we follow the standard evaluation protocol and use the every-8th-image split, where images with indices $0, 8, 16, \ldots$ are reserved for testing.

\subsubsection{Baselines.} We compare MGS against six representative Gaussian LoD baselines.
\textit{Discrete} LoD methods expose a fixed set of quality levels, including
{H3DGS}~\cite{H3DGS-kerbl2024hierarchical} ($\tau$-threshold hierarchy),
{Octree-GS}~\cite{Octree-GS-ren2024octree} (anchor-based octree LOD),
{MaskGaussian}~\cite{MaskGaussian-liu2025maskgaussian} (learnable existence probability), and
{FlexGaussian}~\cite{FlexGaussian-tian2025flexgaussian} (training-free pruning and quantization),
each at the discrete operating points reported in respective papers.
\textit{Continuous} LoD methods support rendering at arbitrary splat budgets, including
{CLoD-GS}~\cite{CLoD-GS-cheng2025clod} (distance-dependent opacity decay) and
{CLoD-3DGS}~\cite{CLoD-3DGS-milef2025learning} (learned importance ordering), are evaluated at the same prefix ratios as MGS: 
100\%, 90\%, $\ldots$, 20\%, 10\%, 5\%, and 1\%.

\setlength{\tabcolsep}{4pt}
\setlength{\fboxrule}{0pt} 
\setlength{\fboxsep}{2pt}
\begin{table*}[t]
\centering
\caption{
Quantitative comparison on four benchmarks.
PSNR, SSIM \cite{SSIM-wang2004image}, LPIPS \cite{LPIPS-zhang2018unreasonable} reported at highest splat budget per baseline.
AUC$_{\text{fps}}$ summarises quality--FPS trade-off frontier, and AUC$_{\text{splats}}$ encapsulates quality/splat efficiency.
The \textbf{best} and \underline{second best} results are highlighted.
$\uparrow$ higher is better, $\downarrow$ lower is better.
}
\label{tab:main_comparison_updated}
\small
\resizebox{\textwidth}{!}{
\begin{tabular}{llccccc|ccccc}  
\toprule
& \multirow{2}{*}{\textbf{Method}} & \multicolumn{5}{c}{\textbf{MipNeRF 360}}
& \multicolumn{5}{c}{\textbf{Tanks \& Temples}} \\
\cmidrule(lr){3-7} \cmidrule(lr){8-12}
& & PSNR$\uparrow$ & SSIM$\uparrow$ & LPIPS$\downarrow$ & AUC$_{\text{fps}}$$\uparrow$ & AUC$_{\text{splats}}$$\uparrow$
& PSNR$\uparrow$ & SSIM$\uparrow$ & LPIPS$\downarrow$ & AUC$_{\text{fps}}$$\uparrow$ & AUC$_{\text{splats}}$$\uparrow$ \\
\midrule
\rowcolor{gray!10}
& {3DGS-MCMC}~\cite{3DGS-MCMC-kheradmand20243d}
& 28.40 & 0.843 & 0.133 & -- & -- & 24.76 & 0.877 & 0.083 & -- & -- \\
\midrule
\multirow{4}{*}{\rotatebox[origin=c]{90}{\textbf{Disc.}}} 
& H3DGS~\cite{H3DGS-kerbl2024hierarchical}        & 25.41 & 0.780 & 0.230 & 11.62 & 60.51 & 20.41 & 0.808 & 0.186 & 8.23 & 58.84  \\
& FlexGaussian~\cite{FlexGaussian-tian2025flexgaussian} & 26.86 & 0.793 & 0.237 & 30.04 & 62.05 & 22.89 & 0.823 & 0.198 & \underline{52.38} & 64.93  \\
& MaskGaussian~\cite{MaskGaussian-liu2025maskgaussian} & 27.42 & \underline{0.815} & 0.218 & 34.11 & 67.10 & 23.66 & 0.846 & 0.181 & 45.15 & 68.89 \\
& Octree-GS~\cite{Octree-GS-ren2024octree}    & \underline{27.62} & 0.813 & 0.221 & 21.61 & \underline{68.43} & \textbf{24.59} & \underline{0.865} & \underline{0.157} & 23.80 & \underline{67.81}  \\
\midrule
\multirow{3}{*}{\rotatebox[origin=c]{90}{\textbf{Cont.}}} 
& CLoD-GS~\cite{CLoD-GS-cheng2025clod}      & 26.55 & 0.803 & 0.233 & 19.39 & 63.43 & 22.78 & 0.832 & 0.193 & 23.39 & 65.37  \\
& CLoD-3DGS~\cite{CLoD-3DGS-milef2025learning}    & 27.44 & 0.814 & \underline{0.215} & \underline{46.38} & 56.56 & 23.51 & 0.845 & 0.176 & 50.97 & 61.89  \\
& \textbf{MGS (Ours)}
& \textbf{28.20} & \textbf{0.841} & \textbf{0.130} & \textbf{64.81} & \textbf{77.79} & \underline{24.56} & \textbf{0.874} & \textbf{0.086} & \textbf{70.16} & \textbf{77.54} \\
\midrule
& \multirow{2}{*}{\textbf{Method}} & \multicolumn{5}{c}{\textbf{Deep Blending}}
& \multicolumn{5}{c}{\textbf{BungeeNeRF}} \\
\cmidrule(lr){3-7} \cmidrule(lr){8-12}
& & PSNR$\uparrow$ & SSIM$\uparrow$ & LPIPS$\downarrow$ & AUC$_{\text{fps}}$$\uparrow$ & AUC$_{\text{splats}}$$\uparrow$
& PSNR$\uparrow$ & SSIM$\uparrow$ & LPIPS$\downarrow$ & AUC$_{\text{fps}}$$\uparrow$ & AUC$_{\text{splats}}$$\uparrow$ \\
\midrule
\rowcolor{gray!10}
& {3DGS-MCMC}~\cite{3DGS-MCMC-kheradmand20243d}
& 27.63 & 0.893 & 0.204 & -- & -- & 27.04 & 0.903 & 0.091 & -- & -- \\
\midrule
\multirow{4}{*}{\rotatebox[origin=c]{90}{\textbf{Disc.}}}
& H3DGS~\cite{H3DGS-kerbl2024hierarchical}        & 27.77 & 0.882 & 0.277 & 10.39 & 73.82 & 27.74 & 0.910 & 0.114 & 8.47 & 62.57 \\
& FlexGaussian~\cite{FlexGaussian-tian2025flexgaussian} & 29.25 & 0.899 & 0.251 & 43.61 & 75.53 & 25.87 & 0.872 & 0.129 & 12.89 & 53.58  \\
& MaskGaussian~\cite{MaskGaussian-liu2025maskgaussian} & \underline{29.69} & \underline{0.907} & \underline{0.245} & 54.96 & 78.77 & \underline{27.80} & \underline{0.917} & 0.099 & 21.58 & 71.18  \\
& Octree-GS~\cite{Octree-GS-ren2024octree}    & \textbf{30.35} & \textbf{0.910} & 0.252 & 29.54 & \underline{83.21} & \textbf{28.23} & \textbf{0.922} & \underline{0.088} & 19.69 & \underline{73.70} \\
\midrule
\multirow{3}{*}{\rotatebox[origin=c]{90}{\textbf{Cont.}}}
& CLoD-GS~\cite{CLoD-GS-cheng2025clod}      & 29.03 & 0.901 & 0.249 & 26.09 & 77.34 & 25.35 & 0.859 & 0.148 & 13.83 & 65.40  \\
& CLoD-3DGS~\cite{CLoD-3DGS-milef2025learning}    & 29.54 & 0.903 & 0.247 & \underline{69.36} & 72.33 & 26.57 & 0.892 & 0.131 & \underline{24.45} & 29.72  \\
& \textbf{MGS (Ours)}
& 28.41 & 0.902 & \textbf{0.176} & \textbf{82.63} & \textbf{83.22}
& 27.13 & 0.906 & \textbf{0.088} & \textbf{62.99} & \textbf{79.93} \\
\bottomrule
\end{tabular}
}
\end{table*}

\subsubsection{Metrics.}
We report image-quality metrics PSNR, SSIM~\cite{SSIM-wang2004image}, LPIPS~\cite{LPIPS-zhang2018unreasonable}.
In addition, a key requirement to evaluating LoD methods is comparing quality--speed trade-off at different operating points.
To this end, we first define a composite \emph{quality} score calculated at each operating point:
\begin{equation}
\label{eq:mixed_quality}
\bar{Q}=\tfrac{1}{3}\bigl(\tilde{p}+\tilde{s}+(1{-}\tilde{\ell})\bigr),
\end{equation}
where $\tilde{p}$, $\tilde{s}$, $\tilde{\ell}$ are PSNR, SSIM, and
LPIPS linearly clamped to $[0,1]$ using fixed ranges
$(14,32)$, $(0.35,0.92)$, and $(0.06,0.60)$, respectively.
We then summarise the full quality--speed trade-off with two
area-under-the-curve (AUC) scores.

\textbf{$\mathrm{AUC}_{\text{fps}}$}($\uparrow$): we construct a monotone envelope of quality versus FPS.
Noting that an operating point can be replicated at any lower speed, the envelope extends leftward from every achieved operating point along lowering throughput.
We clip FPS to $[0,\,500]$ and compute the normalised area under this envelope.

\textbf{$\mathrm{AUC}_{\text{splat}}$}($\uparrow$):  we construct a monotone envelope of quality vs splat count.
Noting that any operating point can be replicated with additional splats, the envelope extends rightward from all operating points along increasing splat count.
Also noting that quality diminishes as splat count approaches $0$, we connect the origin $(0,0)$ to the lowest-budget operating point.
We clip budget to $[0,\,5\text{M}]$ and compute the normalised area under this envelope.

Both \textbf{$\mathrm{AUC}_{\text{fps}}$} and \textbf{$\mathrm{AUC}_{\text{splat}}$} are scaled by $100$ for enhanced readability.

\subsubsection{Implementation.}
We implement MGS on \textit{gsplat}~\cite{gsplat-ye2025gsplat} codebase using 3DGS-MCMC~\cite{3DGS-MCMC-kheradmand20243d} training strategy.
Unless otherwise noted, we order splats by opacity in descending order (\cref{eq:ordering}), use equal prefix/full weights ($\gamma{=}1$, \cref{eq:mgs_objective}), set the full-scene capacity to $N{=}5$M (\cref{eq:ordering}), and train for 50\,k iterations.
All experiments are conducted on identical Ubuntu servers with NVIDIA A100 GPU.

\subsection{Results and Comparisons}
\label{sec:results}

\begin{figure*}[t]
    \centering
    \includegraphics[width=1.0\linewidth]{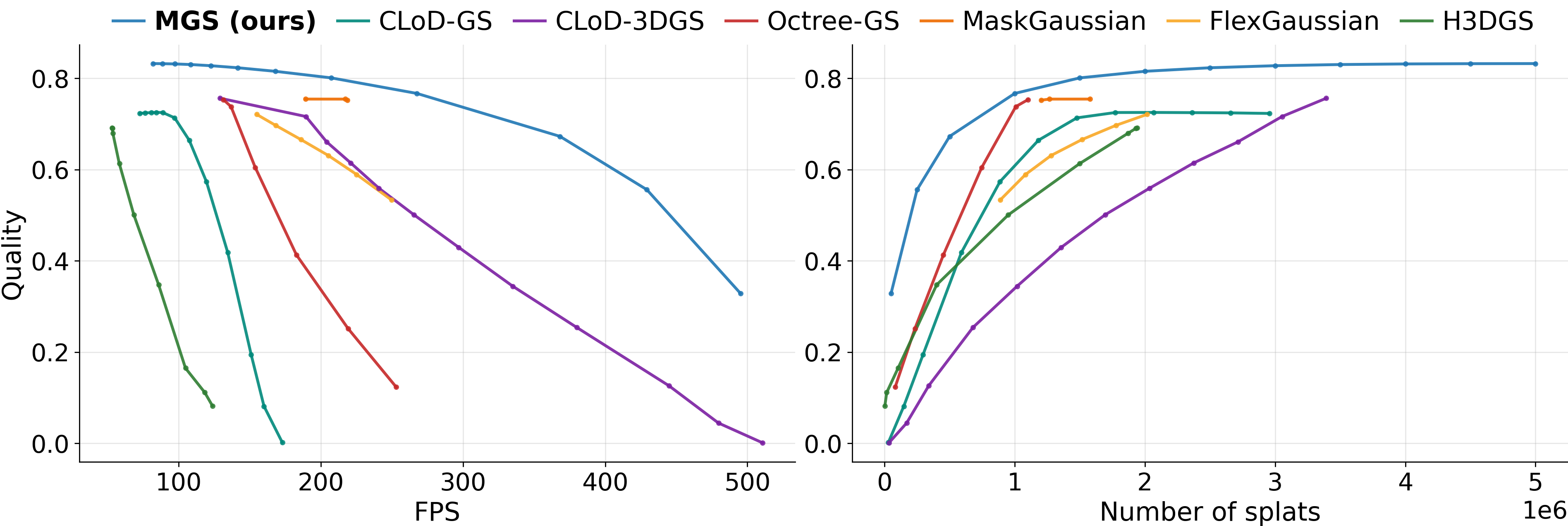}
    \caption{Quality-budget trade-off on Mip-NeRF 360~\cite{MipNeRF360-barron2022mip}, averaged across all nine scenes.
    \textbf{Left:} Quality ($\bar{Q}$, Eq.~\ref{eq:mixed_quality}) vs.\ FPS.
    \textbf{Right:} Quality vs.\ number of Gaussian splats.
    Curves trace continuous LoD models across prefix ratios 1\%--100\%, and trace discrete LoD models at their recommended operating points respectively.
    MGS (ours, dark blue) achieves the highest quality at every speed and splat budget, while spanning a much wider FPS range than any baseline.}
    \label{fig:fps_vs_quality}
    \end{figure*}

\subsubsection{Full-splat Quality Comparison.}
\cref{tab:main_comparison_updated} reports each method at its highest splat-count operating point, averaged per benchmark.
On MipNeRF~360~\cite{MipNeRF360-barron2022mip}, MGS achieves the best PSNR (28.20\,dB), SSIM (0.841), and LPIPS (0.130), outperforming the next-best LoD baseline (Octree-GS~\cite{Octree-GS-ren2024octree}, 27.62\,dB) by +0.58\,dB while achieving substantially lower perceptual error (LPIPS 0.130 vs.\ 0.221).
On Tanks \& Temples~\cite{TanksAndTemples-knapitsch2017tanks}, MGS trails Octree-GS in PSNR by only 0.03\,dB (24.56 vs.\ 24.59) but achieves the best SSIM and LPIPS.
On Deep Blending~\cite{DeepBlending-hedman2018deep} and BungeeNeRF~\cite{BungeeNeRF-xiangli2022bungeenerf}, Octree-GS obtains higher PSNR at its single highest-quality level; however, its coarser LoD levels incur severe quality degradation, yielding far lower AUC scores than MGS.
MGS consistently achieves the lowest LPIPS, indicating strong perceptual quality across all the four benchmarks.

\begin{figure*}[thbp]
\centering
\includegraphics[width=1.0\linewidth]{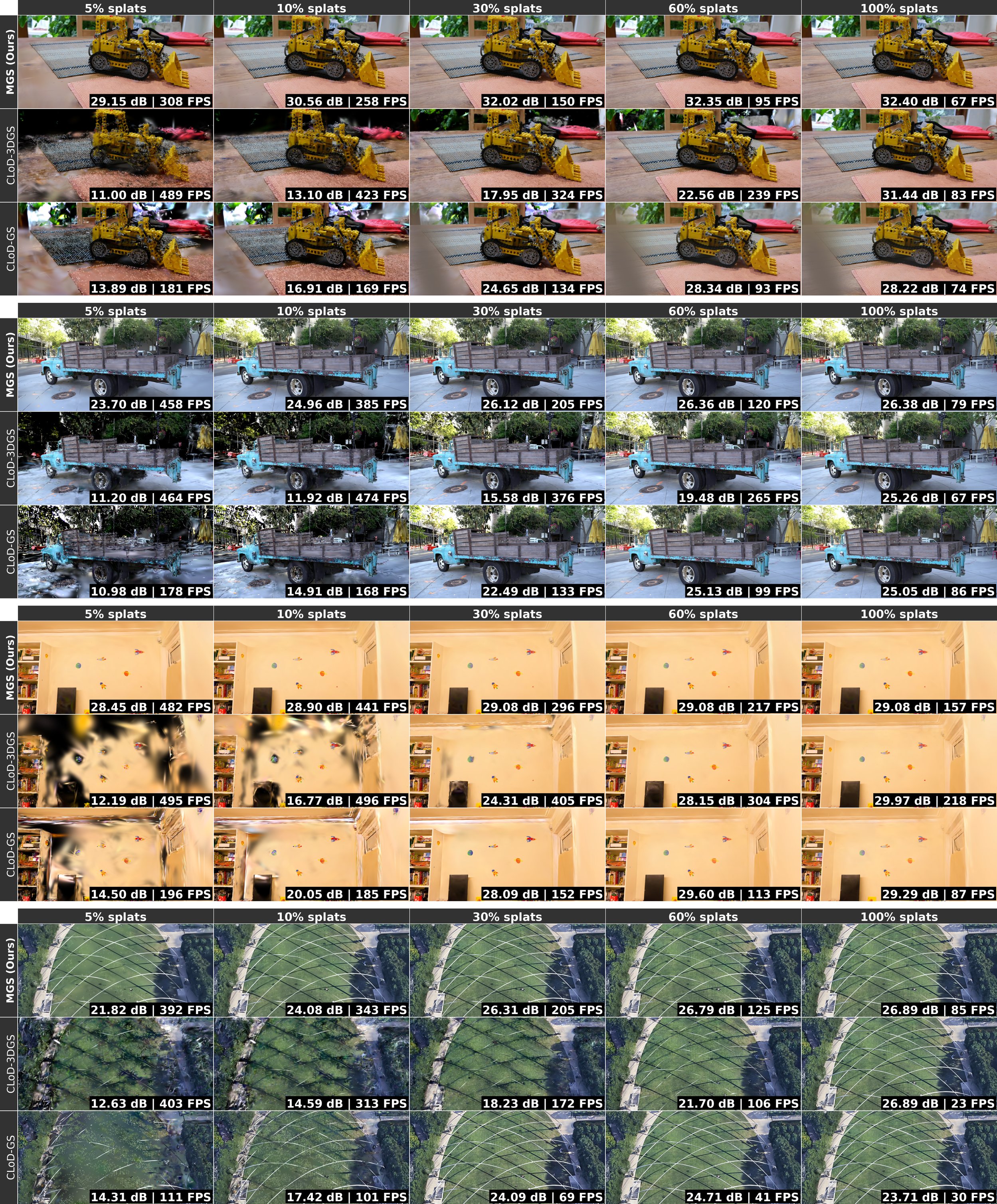}\\
\caption{Qualitative comparison of continuous LoD methods across four benchmarks. Renderings are shown at 5\%, 10\%, 30\%, 60\%, and 100\% of the full splat budget. We compare MGS with CLoD-3DGS~\cite{CLoD-3DGS-milef2025learning} and CLoD-GS~\cite{CLoD-GS-cheng2025clod}. Under highly constrained budgets (5–10\%), MGS maintains coherent reconstructions with PSNR of 21--28\,dB, while both baselines suffer from severe artifacts and quality collapse (11--17\,dB).
}
\label{fig:visual_comparison_1}
\end{figure*}

\begin{figure*}[thbp]
\centering
\includegraphics[width=1.0\linewidth]{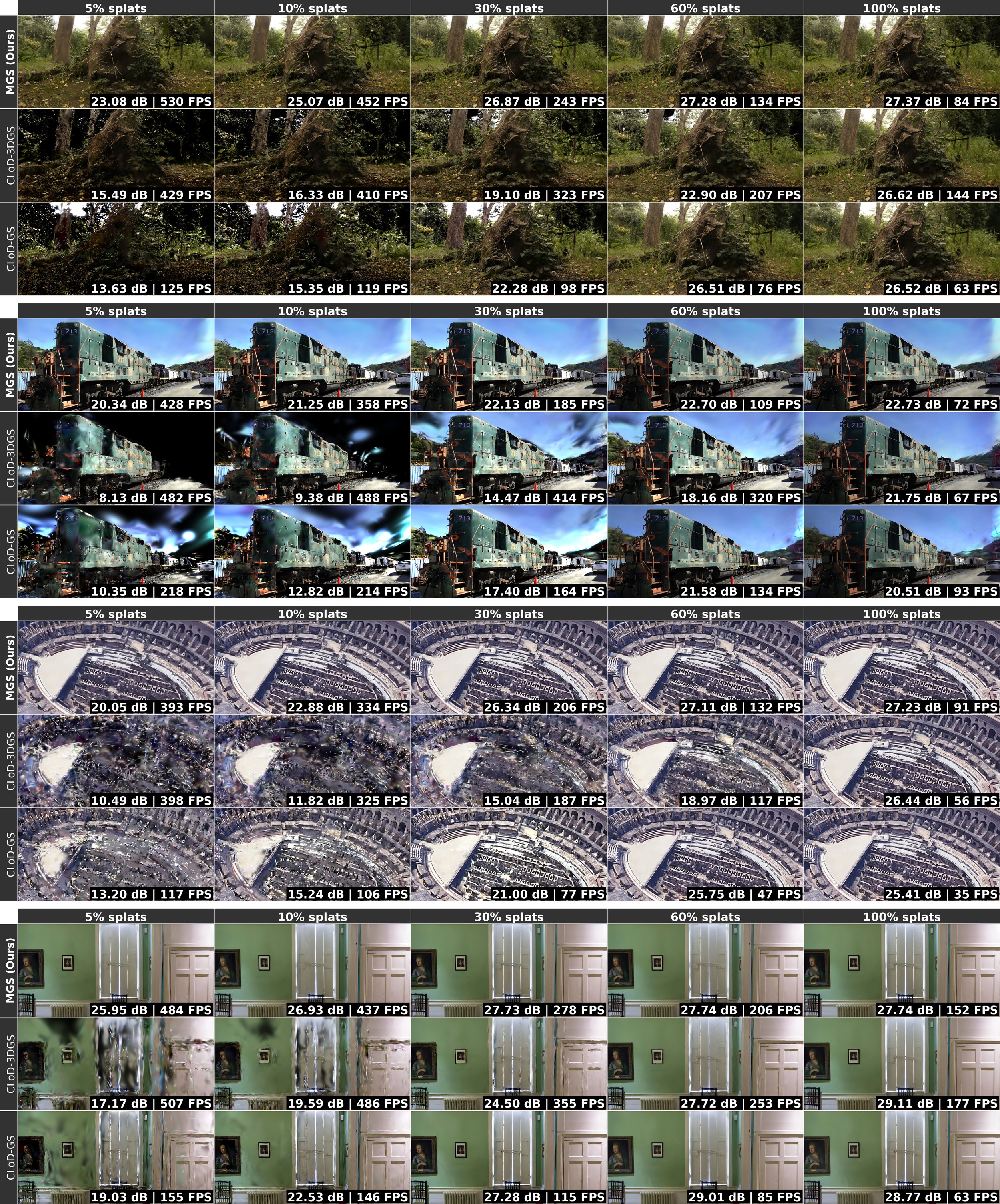}\\
\caption{Additional qualitative results including a failure case.
On stump, train, and rome, MGS maintains higher fidelity across all budget levels.
On DrJohnson (bottom), CLoD-3DGS~\cite{CLoD-3DGS-milef2025learning} achieves higher peak PSNR at 100\% budget (29.1 vs.\ 27.7\,dB);
however, MGS still degrades more gracefully at reduced budgets (5--30\%).}
\label{fig:visual_comparison_2}
\end{figure*}

\subsubsection{Quality--Speed Trade-off.}
MGS outperforms all baselines in both AUC${\text{fps}}$ and AUC${\text{splats}}$ by wide margins across four benchmarks, sustaining high fidelity consistently across varying speed and budget constraints (\cref{fig:fps_vs_quality}).
Qualitative comparisons also confirm that MGS preserves coherent scene structure at aggressive budget reductions (5--10\% splats) \cref{fig:visual_comparison_1,fig:visual_comparison_2}, whereas CLoD-3DGS and CLoD-GS suffer from severe artifacts; the one exception is DrJohnson, where CLoD-3DGS achieves higher peak PSNR at full budget yet degrades more sharply at reduced budgets.

By adjusting the prefix ratio from 1\% to 100\%, MGS produces a smooth, dense Pareto frontier of operating points from a single trained model, without any per-budget retraining or model switching.
Among continuous-LoD competitors, CLoD-3DGS~\cite{CLoD-3DGS-milef2025learning} achieves the next-highest AUC$_{\text{fps}}$ on MipNeRF~360 (28.94 vs.\ 54.46 for MGS) but at substantially lower image quality (27.44 vs.\ 28.20\,dB PSNR, 0.215 vs.\ 0.130 LPIPS).
CLoD-GS~\cite{CLoD-GS-cheng2025clod} spans a narrower speed range, resulting in lower AUC$_{\text{fps}}$ (9.78 on MipNeRF~360).
Discrete-LoD methods such as Octree-GS~\cite{Octree-GS-ren2024octree} and H3DGS~\cite{H3DGS-kerbl2024hierarchical} exhibit low AUC scores (AUC$_{\text{fps}}$: 9.96 and 4.81 on MipNeRF~360), because their coarser levels incur severe quality drops.
FlexGaussian~\cite{FlexGaussian-tian2025flexgaussian} offers competitive quality through training-free compression but produces only a handful of discrete operating points and achieves lower AUC values than MGS across all benchmarks.

\subsubsection{Continuous vs.\ Discrete Operating Points.}
A key practical advantage of MGS is the \emph{density} of the operating-point frontier.
MGS produces a coherent rendering for every integer splat budget $k\in\{1,\ldots,N\}$; in practice, evaluating 12 budget ratios already yields a smooth quality--speed curve (\cref{fig:fps_vs_quality}).
In contrast, Octree-GS~\cite{Octree-GS-ren2024octree} provides 3--6 LOD levels, H3DGS~\cite{H3DGS-kerbl2024hierarchical} offers 9 $\tau$-thresholds, and FlexGaussian~\cite{FlexGaussian-tian2025flexgaussian} exposes 2--6 compression targets, each requiring separate configurations.
This distinction matters for deployment: MGS allows a system to respond to per-frame or per-device budgets by simply truncating the splat array, with no additional data structures, mode switches, or latency spikes.

\subsubsection{Quality vs 3DGS-MCMC.}
Although MGS trains for \emph{all} possible splat budgets, its full-quality performance closely approaches and sometimes exceeds that of stand-alone 3DGS-MCMC~\cite{3DGS-MCMC-kheradmand20243d} backbone, which trains only for the full set and provides no LoD capability.
On MipNeRF~360 and Tanks \& Temples, MGS trails 3DGS-MCMC by only 0.20\,dB (28.20 vs.\ 28.40 and 24.56 vs.\ 24.76, respectively).
On Deep Blending and BungeeNeRF, MGS \emph{surpasses} 3DGS-MCMC (28.41 vs.\ 27.63\,dB and 27.13 vs.\ 27.04\,dB), suggesting that the stochastic budget objective can act as a beneficial regulariser on certain scene types.
This confirms that continuous LoD need not come at the cost of full-capacity quality.

\begin{figure}[t]
\centering
\includegraphics[width=1.0\linewidth]{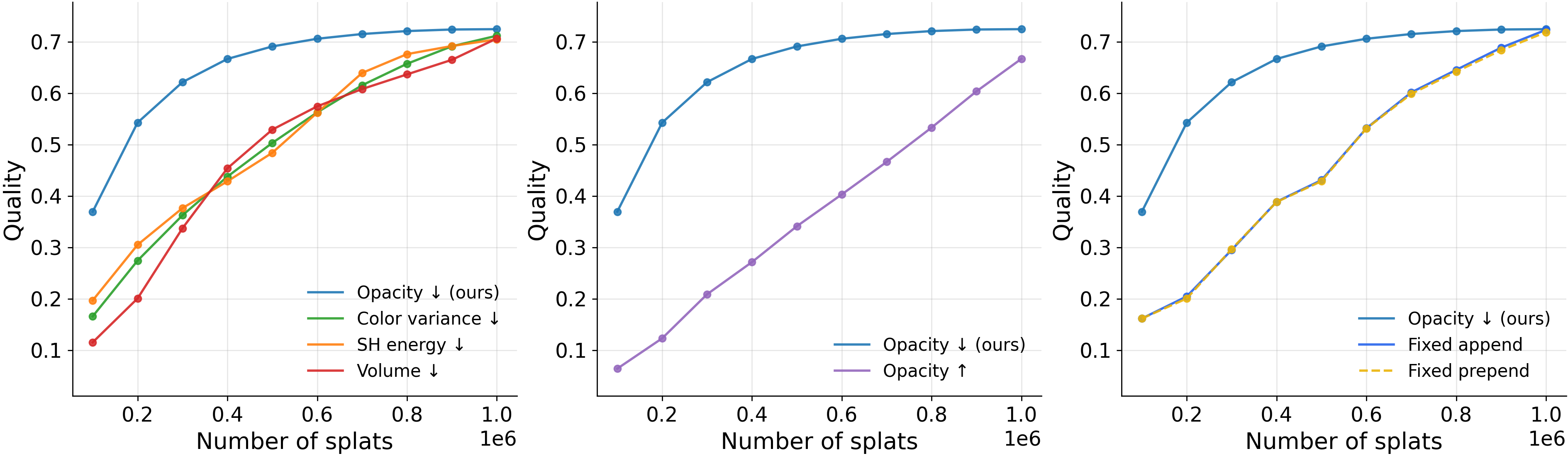}\
\caption{Ablations on the importance score.
The opacity-descending dominates other importance scores across the full budget range.}
\label{fig:ablation_ordering_strategy}
\end{figure}

\subsection{Ablation Studies}
\label{sec:ablations}

We conduct controlled ablations on a single scene (bicycle from MipNeRF 360~\cite{MipNeRF360-barron2022mip}) with $N{=}1$M splats and 50\,k training steps unless otherwise noted, isolating each design choice while keeping the others at their default values \cref{tab:ablation}.

\subsubsection{Importance Score.}
We evaluate seven importance scores derived from four per-splat criteria $s(g)$, \ie~opacity, volume, SH energy, and colour variance (\cref{eq:ordering}), sorting in ascending and descending order, as well as sorting criteria not determined by splat characteristics  (\cref{fig:ablation_ordering_strategy}). 
Sorting by opacity in descending order consistently produces the most effective multi-budget performance among all strategies: at 10\% of the splat budget, it achieves 22.2\,dB PSNR at 493\,FPS, whereas the next-best score-based ordering (SH-energy descending) reaches only 17.6\,dB under the same constraints.
Sorting in ascending order performs poorly overall, demonstrating worse performance than the descending variant.

We compare against two fixed insertion orders: \texttt{append} and \texttt{prepend} (\cf~\cref{sec:method}).
The two fixed insertion orders underperform opacity-descending at low prefix ratios: for example, at 10\% budget, fixed-random reaches 21.5\,dB  compared to 22.2\,dB for opacity-descending. This confirms the importance of a semantically meaningful coarse-to-fine ordering for preserving early-prefix quality,
yielding 10.5\,dB and 11.5\,dB,, respectively, at the 10\% of the splat budget.

\setlength{\tabcolsep}{4pt}
\setlength{\fboxrule}{0pt}
\setlength{\fboxsep}{2pt}
\begin{table*}[t!]
\centering
\caption{
Ablation study on importance score and budget training strategy.
All metrics reported at split=1.0 (full budget).
PSNR, SSIM and LPIPS represent raw quality; AUC$_{\text{fps}}$ and AUC$_{\text{splats}}$ summarise quality--efficiency frontiers.
The \textbf{best} and \underline{second best} results per column are highlighted. $\uparrow$ higher is better, $\downarrow$ lower is better.
}
\label{tab:ablation}
\small
\resizebox{\textwidth}{!}{
\begin{tabular}{llccccc}
\toprule
& & \textbf{PSNR$\uparrow$} & \textbf{SSIM$\uparrow$} & \textbf{LPIPS$\downarrow$} & \textbf{AUC$_{\text{fps}}$$\uparrow$} & \textbf{AUC$_{\text{splats}}$$\uparrow$} \\
\midrule
\multirow{7}{*}{Importance Score}
& \textbf{Opacity $\downarrow$ (ours)} & \underline{25.47} & \textbf{0.776} & \textbf{0.174} & \textbf{68.62} & \textbf{61.05} \\
& Opacity $\uparrow$ & 24.91 & 0.743 & 0.219 & 30.53 & 33.48 \\
\cmidrule(lr){2-7}
& Color variance $\downarrow$ & 25.25 & 0.770 & 0.183 & 51.94 & 46.26 \\
& SH energy $\downarrow$ & 25.17 & 0.766 & 0.187 & \underline{59.39} & \underline{47.12} \\
& Volume $\downarrow$ & 25.19 & 0.764 & 0.183 & 52.15 & 44.73 \\
\cmidrule(lr){2-7}
& Fixed append & \textbf{25.51} & \underline{0.773} & \underline{0.175} & 46.41 & 43.09 \\
& Fixed prepend & 25.32 & 0.772 & 0.176 & 49.65 & 42.90 \\
\midrule
\multirow{3}{*}{Budget Training}
& \textbf{Prefix + full (ours)} & \textbf{25.43} & \textbf{0.776} & \textbf{0.175} & \textbf{66.87} & \textbf{60.96} \\
& Prefix & 24.97 & 0.734 & 0.253 & 56.58 & 57.09 \\
& MRL & \underline{25.37} & \underline{0.771} & \underline{0.183} & \underline{66.57} & \underline{60.35} \\
\bottomrule
\end{tabular}
}
\end{table*}

\begin{figure}[t!]
\centering
\includegraphics[width=1.0\linewidth]{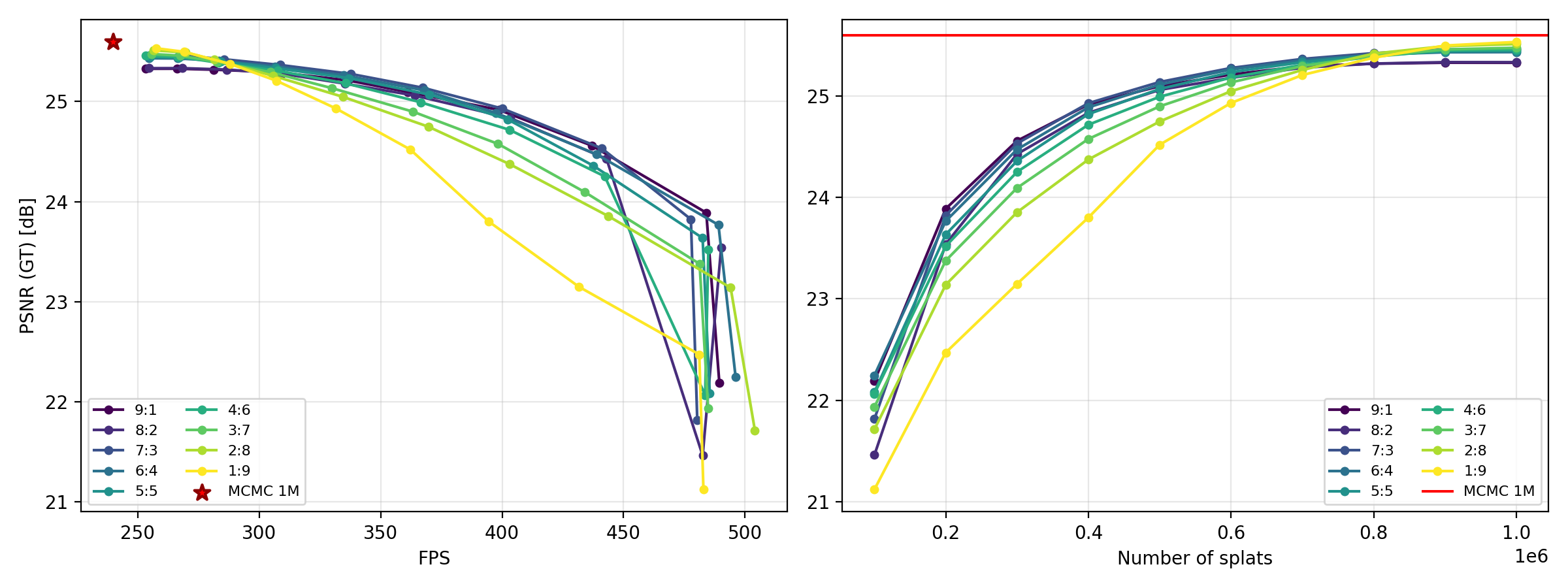}\
\caption{Ablations on Prefix/Full Loss Weight distribution.
While prefix-heavy or full-heavy weighting can marginally improve their respective endpoints, a 6:4 ratio provides the best area under the curve.
For the sake of simplicity and balanced performance, we utilise a 5:5 ratio for  our MGS approach and for all other ablation experiments.
}
\label{fig:ablation_loss_weight}
\end{figure}

\subsubsection{Budget Training.}
Fixing our opacity-descending importance score, we compare three budget training strategies:
\texttt{prefix + full} (our default; \cref{eq:mgs_objective}),
\texttt{prefix only} (no full-set term),
and \texttt{MRL nesting}~\cite{MRL-kusupati2022matryoshka} (deterministic, fixed prefix sizes).
The \texttt{prefix + full} achieves the best trade-off between low-budget and full-quality performance.
Removing the full-set term (\texttt{prefix only}) improves early prefixes marginally but degrades quality at higher budgets.
\texttt{MRL nesting} restricts the model to a discrete set of budget sizes, sacrificing the fine-grained coverage that stochastic sampling provides.

\subsubsection{Prefix/Full Loss Weight.} %
Sweeping the prefix\,:\,full weight ratio from 9:1 to 1:9 reveals a clear trade-off (\cref{fig:ablation_loss_weight}).
At the 10\% prefix, prefix-heavy weighting (9:1) attains 22.2\,dB whereas the full-heavy setting (1:9) drops to 21.1\,dB.
Conversely, at full quality, the 1:9 ratio achieves the best performance (PSNR 25.5\,dB and LPIPS 0.169), surpassing 9:1 by +0.2\,dB.
For simplicity, we use \textbf{1:1} (5:5) as the default setting, as it achieves comparable performance (25.4\,dB at full quality and 22.1\,dB at 10\% prefix quality).

\section{Conclusion}
\label{sec:conclusion}

In this work, we present Matryoshka Gaussian Splatting (MGS), a continuous budget control framework for 3DGS models.
By learning an ordered, prefix-closed set of Gaussian primitives, MGS allows rendering at arbitrary splat budgets by truncating the primitive set, producing a dense spectrum of quality--speed operating points.
To train a single representation to perform well at every prefix length, we introduce a stochastic budget training strategy, which can be efficiently integrated into existing 3DGS pipelines without architectural changes.
Experiments on four standard benchmarks demonstrate that MGS matches the full-capacity performance of 3DGS while enabling continuous LoD control from a single model.
Extensive ablations further confirm the effectiveness of the proposed ordering strategy, training objective, and budget sampling designs.
Our results suggest that nested primitive representations provide a promising direction for scalable neural scene representations.
Future work could explore distance or view-dependent prefix selection, adaptive budget scheduling, and integration with streaming or device-aware rendering systems.

\makeatletter
\ifeccv@review
\else
\section*{Acknowledgements}
This work was supported by a UKRI Future Leaders Fellowship [grant number G127262].
\fi
\makeatother

\clearpage

\bibliographystyle{splncs04}
\bibliography{reference}

\end{document}